# Image Pixel Fusion for Human Face Recognition

Mrinal Kanti Bhowmik[1], Debotosh Bhattacharjee[2], Mita Nasipuri[2], Dipak Kumar Basu[2*], and Mahantapas Kundu[2]

[1]Department of Computer Science and Engineering, Tripura University
Suryamaninagar- 799130, Tripura, India
Email: mkb_cse@yahoo.co.in

[2]Department of Computer Science and Engineering, Jadavpur University
Kolkata- 700032, India
*AICTE Emeritus Fellow
Email: debotosh@indiatimes.com, {mita_nasipuri, dipakkbasu}@gmail.com , mkundu@icse.jdvu.ac.in

*Abstract*—In this paper we present a technique for fusion of optical and thermal face images based on image pixel fusion approach. Out of several factors, which affect face recognition performance in case of visual images, illumination changes are a significant factor that needs to be addressed. Thermal images are better in handling illumination conditions but not very consistent in capturing texture details of the faces. Other factors like sunglasses, beard, moustache etc also play active role in adding complicacies to the recognition process. Fusion of thermal and visual images is a solution to overcome the drawbacks present in the individual thermal and visual face images. Here fused images are projected into an eigenspace and the projected images are classified using a radial basis function (RBF) neural network and also by a multi-layer perceptron (MLP). In the experiments Object Tracking and Classification Beyond Visible Spectrum (OTCBVS) database benchmark for thermal and visual face images have been used. Comparison of experimental results show that the proposed approach performs significantly well in recognizing face images with a success rate of 96% and 95.07% for RBF Neural Network and MLP respectively.

## I. INTRODUCTION

Human face recognition has already established its acceptance as a superior biometric method for identification and authentication purposes. It is touch less, highly automated and most natural since it coincides with the mode of recognition that we as humans employ on our everyday affairs [1]. It has emerged as a preferred alternative to traditional forms of identification, like card IDs, which are not embedded into one's physical characteristics. Research into several biometric modalities including face, fingerprint, iris, and retina recognition has produced varying degrees of success [2]. It has many practical applications, such as bankcard identification, access control, mug shots searching, security monitoring, surveillance systems etc [3], [4], [5].

Most of the research efforts in this area have focused on visible spectrum imaging and geometric feature extraction. Despite the success of automatic face recognition techniques in many practical applications, the task of face recognition based only on the visible spectrum is still a challenging problem under uncontrolled environments. The challenges are even more profound when one considers the large variations in the visual stimulus due to illumination conditions, viewing directions or poses, facial expressions, aging, and disguises such as facial hair, glasses, or cosmetics. In this connection, there are two major challenges: variations in illumination and pose [6]. Such problems are quite unavoidable in applications such as outdoor access control and surveillance. Performance of visual face recognition is sensitive to variations in illumination conditions and usually degrades significantly when the lighting is dim or when it is not uniformly illuminating the face. The changes caused by illumination on the same individual are often larger than the differences between individuals. Various algorithms (e.g. histogram equalization, dropping leading eigenfaces etc.) for compensating such variations have been studied with partial success. These techniques attempt to reduce the within-class variability introduced by changes in illumination. A visual face recognition system optimized for identification of light-skinned people could be prone to higher false alarms among dark-skinned people.

Thermal IR imagery [7] has been suggested as a viable alternative in detecting disguised faces and handling situations where there is no control over illumination. Thermal IR images represent the heat patterns emitted from an object. Objects emit different amounts of IR energy according to their body temperature and characteristics. Since, vessels transport warm blood throughout the body; the thermal patterns of faces are derived primarily from the pattern of blood vessels under the skin. The vein and tissue structure of the face is unique for each person, and therefore the IR images are also unique. It is known that even identical twins have different thermal patterns. Face recognition based on thermal IR spectrum utilizes the anatomical information of human face as features unique to each individual while sacrificing color recognition. Anatomical features of faces useful for identification can be measured at a distance using passive IR sensor technology with or without the cooperation of the subject [6].

Recently, researchers have investigated the use data fusion method which combines different types of data gathered by the simultaneous use of several sensing modalities to generate a new type of data. Various perceptual mechanisms integrate these senses to produce the internal representation of the sensed environment. The integration tends to be synergistic in the scene that information inferred from the process cannot be obtained from any proper subset of the sense modalities. This property of synergism is one that should be sought for when implementing multisensor integration for machine perception. The principal motivation for the fusion approach is to exploit such synergism in the technique for combined interpretation of images obtained from multiple sensors.

Research work on fusion has been carried out only for last few years. Fusion methods can be classified into two categories as stated in [8]. One is about weakly coupled fusion methods, and the other is about strongly coupled fusion methods. In the first category of fusion methods, fusion of data produced by sensory modules does not affect the operation of the modules. On the contrary, for strongly coupled fusion methods, the modules producing the data to be fused are being affected in some way by other information from other modules.

The detailed review on current advances in visual and thermal face recognition is given in [9]. Simple image fusion in spatial domain is discussed in [10], where face recognition is used for testing the fusion of face database images. Both image fusion and decision fusion is employed in [11] to improve the accuracy of the face recognition system.

In this paper, we present a new approach to the problem of face recognition that realizes the full potential of fusion of thermal IR band and visual band images. In this work at first thermal and visual face images are combined together and fused image of corresponding thermal and visual face images are obtained. After that using these transformed fused images eigenfaces are computed and finally those eigenfaces thus found are classified using a radial basis function neural network in the first case and in the second case the eigenfaces are classified using multilayer perceptron.

The organization of the rest of this paper is as follows. In section II, the overview of the system is discussed, in section III experimental results and discussions are given. Finally, section IV concludes this work.

## II. THE SYSTEM OVERVIEW

In this work we have used Object Tracking and Classification Beyond Visible Spectrum (OTCBVS) database benchmark thermal and visual face images. Every thermal face image and the corresponding visual face image are first combined and converted into fused image. These transformed images are separated into two groups namely training and testing set. The eigenspace is computed using training images. All the training and testing images are projected into the created eigenspace and named as fused eigenfaces. After all the conversions a classifier is used to classify them. In first case a radial basis function neural network and in second case a multilayer perceptron are used for this purpose. The block diagram of the system is given in Fig.1. In this figure dotted lines indicate feedback from different steps to their previous steps to improve the efficiency of the systems.

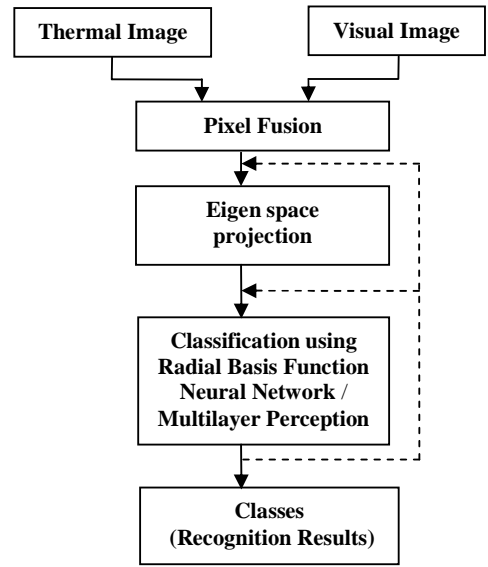

Fig. 1: Block diagram of the present system.

### A. Thermal Infrared Face Images

Thermal infrared face images are formed as a map of the major blood vessels present in the face. Therefore, a face recognition system designed based on thermal infrared face images cannot be evaded or fooled by forgery, or disguise, as can occur using the visible spectrum for facial recognition. Compared to visual face-recognition systems this recognition system will be less vulnerable to varying conditions, such as head angle, expression, or lighting.

### B. Image Fusion Technique

The task of interpreting images, either visual images alone or thermal images alone, is an unconstraint problem. The thermal image can at best yield estimates of surface temperature that, in general, is not specific in distinguishing between object classes. The features extracted from visual intensity images also lack the specificity required for uniquely determining the identity of the imaged object.

The interpretation of each type of image thus leads to ambiguous inferences about the nature of the objects in the scene. The use of thermal data gathered by an infrared camera, along with the visual image, is seen as a way of resolving some of these ambiguities. On the other hand, thermal images are obtained by sensing radiation in the infrared spectrum. The radiation sensed is either emitted by an object at a non-zero absolute temperature, or reflected by it. The mechanisms that produce thermal and visual images are different from each other. Thermal image produced by an object's surface can be interpreted to identify these mechanisms. Thus, thermal images can provide information about the object being imaged which is not available from a visual image [8].

A great deal of effort has been expended on automated scene analysis using visual images, and some work has been done in recognizing objects in a scene using infrared images. However, there has been little effort on interpreting thermal images of outdoor scenes based on a study of the mechanism that gives rise to the differences in the thermal behavior of object surfaces in the scene. Also, nor has been any effort been made to integrate information extracted from the two modalities of imaging.

In our method the process of image fusion is where pixel data of 70% of visual image and 30% of thermal image of same class or same image is brought together into a common operating image or now commonly referred to as a Common Relevant Operating Picture (CROP) [12]. This implies that an additional degree of filtering and intelligence is to be applied to the pixel streams to present pertinent information to the user. So image pixel fusion has the capacity to enable seamless working in a heterogeneous work environment with more complex data. For accurate and effective face recognition we require more informative images. Image by one source (i.e. thermal) may lack some information which might be available in images by other source (i.e. visual). So if it becomes possible to combine the features of both the visual and thermal face images then efficient, robust, and accurate face recognition can be developed.

We describe below in detail the fusion scheme considered in this work. We assume that each face is represented by a pair of images, one in the IR spectrum and one in the visible spectrum. Both images have been combined prior to fusion to ensure similar ranges of values.

We fused visual and thermal images. Ideally, the fusion of common pixels can be done by pixel-wise weighted summation of visual and thermal images [9], as below.

$$F(x, y) = a(x, y)V(x, y) + b(x, y)T(x, y) \quad (1)$$

where $F(x, y)$ is a fused output of a visual image, $V(x, y)$, and a thermal image, $T(x, y)$, while $a(x, y)$ and $b(x, y)$ represent the weighting factors for visual and thermal images respectively. In this work, we have considered $a(x, y) = 0.70$ and $b(x, y) = 0.30$.

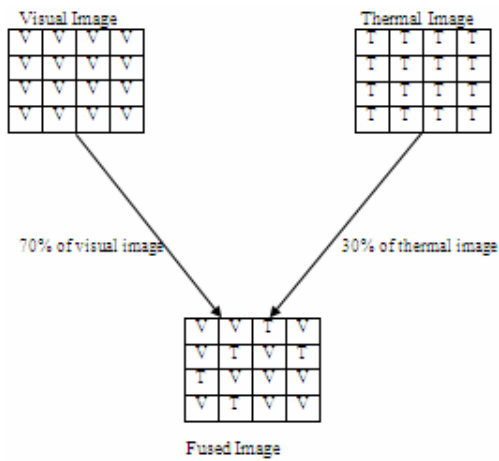

Fig. 2: Fusion Technique.

### C. Eigenfaces for Recognition

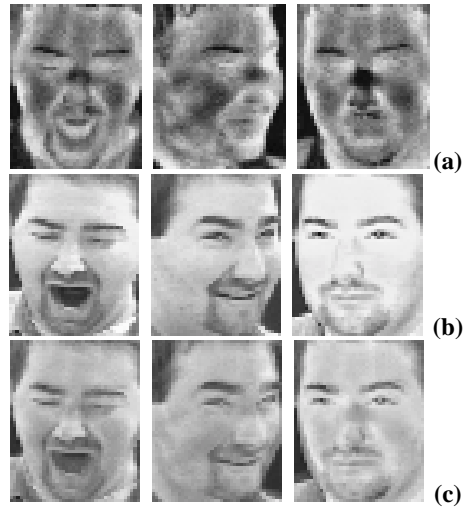

Fig. 3: (a) Thermal Images, (b) Visual Images, (c) Fused Images of corresponding thermal and visual images.

In mathematical terms, we wish to find principal components [13], [14], [15] of the distribution of faces, or the eigenvectors of the covariance matrix of the set of face images.

These eigenvectors can be thought of as set of features which together characterize the variations between face images. Each image location contributes more or less to each eigenvector, so that we can display the eigenvector as sort of ghostly face which we call an eigenface. Each face image in the training set can be presented exactly in terms of a linear combination of the eigenfaces. The number of a possible eigenfaces is equal to the number of face images in the training set. However the faces can also be approximated using only the "best" eigenfaces, those that have the largest eigenvalues and which therefore account for the most variance within the set face images. The best U eigenfaces constitute a U-dimensional subspace, which may be called as "face space" of all possible images. Identifying images through eigenspace projection takes three basic steps. First the eigenspace must be created using training images. After that all those training images are projected into the eigenspace and call them eigenfaces. Train a classifier using these eigenfaces. Finally, the test images are identified by projecting them into the eigenspace and classifying them by the trained classifier.

### D. ANN Using Back propagation with Momentum

Neural networks, with their remarkable ability to derive meaning from complicated or imprecise data, can be used to extract patterns and detect trends that are too complex to be noticed by either humans or other computer techniques. A trained neural network can be thought of as an "expert" in the category of information it has been given to analyze. The Back propagation learning algorithm is one of the most historical developments in Neural Networks. It has reawakened the scientific and engineering community to the modeling and processing of many quantitative phenomena using neural networks. This learning algorithm is applied to multilayer feed forward networks consisting of processing elements with continuous differentiable activation functions. Such networks associated with the back propagation learning algorithm are also called back propagation networks.

*E. Classification of Fused Eigenfaces using Radial Basis Function Network [16]*

Neural networks have been employed and compared to conventional classifiers for a number of classification problems. The results have shown that the accuracy of the neural network approaches is equivalent to or slightly better than other methods. Also, due to the simplicity, generality and good learning ability of the neural networks, these types of classifiers are found to be more efficient.

Radial Basis Function (RBF) neural networks are found to be very attractive for many engineering problems because (1) they are universal approximates, (2) they have a very compact topology and (3) their learning speed is very fast because of their locally tuned neurons. An important property of RBF neural networks is that they form a unifying link between many different research fields such as function approximation, regularization, noisy interpolation and pattern recognition. Therefore, RBF neural networks serve as an excellent candidate for pattern classification where attempts have been carried out to make the learning process in this type of classification faster than normally required for the multilayer feed forward neural networks [17].

In this paper, an RBF neural network is used as a classifier in a face recognition system where the inputs to the neural network are feature vectors derived from the proposed feature extraction technique described in II-B.

Geometrically, the key idea of an RBF neural network is to partition the input space into a number of subspaces which are in the form of hyperspheres. Accordingly, clustering algorithms (k-means clustering, fuzzy k-means clustering and hierarchical clustering) which are widely used in RBF neural networks [18], [19] are a logical approaches to initial centers[18], [20]. However, it may be noted that these clustering approaches are inherently unsupervised learning algorithms as no category information about patterns is used. As an illustrative example, consider a simple training set $(x_k, y_k)$ illustrated in Fig. 4. The black and white data points reflect the corresponding values assumed by the dependent variable $y_k$. If we simply use $k$-means clustering approach without considering $y_k$, two evident clusters as shown in Fig. 4 (a) are achieved. This brings about significant misclassification initially. Although the clustering boundaries are modified in the subsequent learning phase, this could easily lead to an undesired and highly dominant averaging phenomenon as well as to make the learning less effective [19]. To preserve homogeneous clusters, three clusters as depicted in Fig. 4(b) should be created. In other words, a supervised clustering procedure which takes into consideration the category information of training data should be considered.

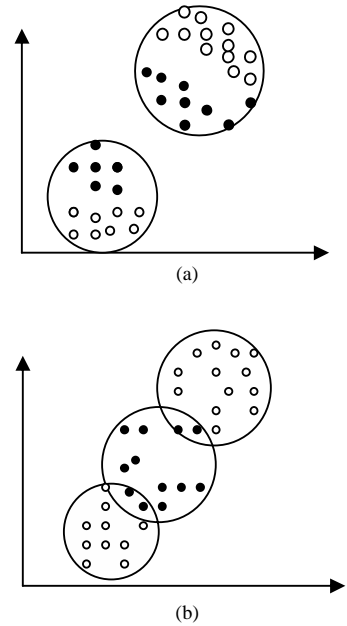

Fig. 4: Two-dimensional patterns and clustering:
(a) conventional clustering,
(b) Clustering with homogeneous analysis.

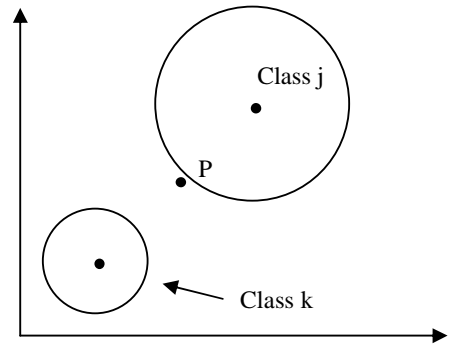

Fig. 5: Effect of Gaussian widths in clustering.

While considering the category information of training patterns, it should be emphasized that the class memberships are not only depended on the distance of patterns, but also depended on the Gaussian widths. As illustrated in Fig. 5, $P$ is near to the center of class $k$ in Euclidean distance, but we can select different Gaussian widths for each cluster so that the point $p$ has greater class membership to class $j$ than that to class $k$. Therefore, the use of class membership implies that we should propose a supervised procedure to cluster the training patterns and determine the initial Gaussian widths.

### III. EXPERIMENTAL RESULTS AND DISCUSSIONS

This work has been simulated using MATLAB 7. For comparison of results experiments are conducted for fused images. A thorough system performance investigation, which covers all conditions of human face recognition, has been conducted. They are face recognition under i) variations in size, ii) variations in lighting conditions, iii) variations in facial expressions, iv) variations in pose.

© ACADEMY PUBLISHER

We first analyze the performance of our algorithm using OTCBVS database which is a standard benchmark thermal and visual face images for face recognition technologies.

*A. OTCBVS database*

Our experiments were perform on the face database which is Object Tracking and Classification Beyond Visible spectrum (OTCBVS) benchmark database contains a set of thermal and visual face images. There are 700 images of visual and 700 thermal images of 16 different persons. For some subject, the images were taken at different times which contain quite a high degree of variability in lighting, facial expression (open/closed eyes, smiling/non smiling etc.), pose (Up right, frontal position etc.) and facial details (Glasses/no Glasses). All the images were taken against a dark homogeneous background with the subjects in and upright, fontal position, with tolerance for some tilting and rotation of up to 20 degree. The variation in scale is up to about 10% all the images in the database.

*B. Classification of Fused Eigenfaces using Radial Basis Function Neural Network and Multilayer Perceptron*

Out of total 700 thermal and visual images 400 images are taken out of which 200 are thermal images and 200 are visual images. Combining these thermal and visual images we get 200 fused images. 100 of these images are used as training set and rest 100 images are taken as testing images. The training set contains 10 classes which mean that each class has 10 images. Now 5 images from one particular class (which are not used as a training image) and 5 more images of the other classes are taken from the testing set. According to this process for all the 10 classes we get the results for both the cases i.e., for classification of fused eigenfaces using RBF Neural Network and using Multilayer Perceptron Neural Network which are shown in Fig. 6 below.

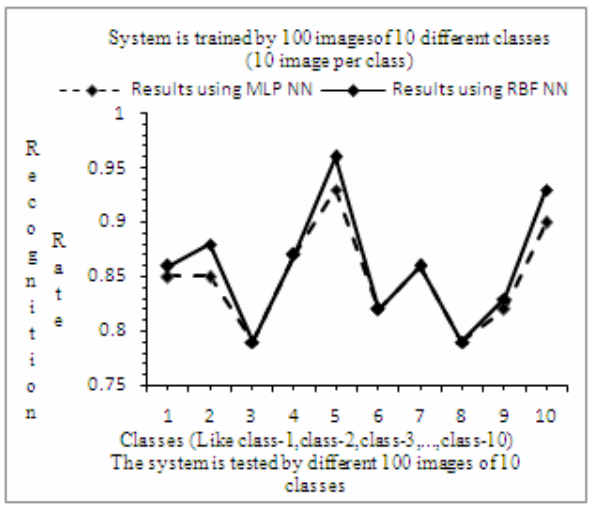

Fig. 6: Comparison between RBF and MLP Classifiers.

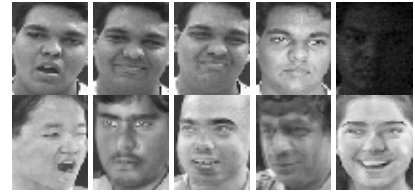

Fig. 7: 10 numbers of fused images used as the testing set of class-1 (which is not used in training).

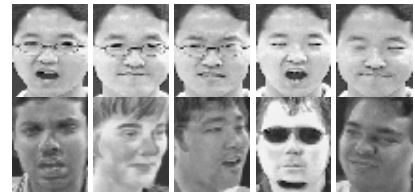

Fig. 8: 10 numbers of fused images used as the testing set of class-2 (which is not used in training).

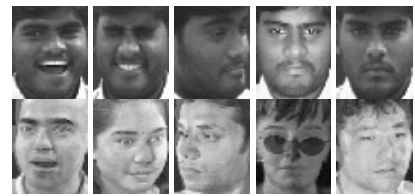

Fig. 9: 10 numbers of fused images used as the testing set of class-3 (which is not used in training).

In the graph shown in Fig. 6, the solid line shows the results of experiments using RBF NN and the dashed line shows the results of experiments using MLP NN. So, from the graph we can easily say that RBF Neural Network gives better result than MLP Neural Network.

In the above figures (from Fig. 7 to Fig. 9) we have shown the fused images which are used in the testing set of Class 1, 2 & 3. These images are of different expressions, variations and different conditions such as lightening, darkness etc.

## IV. CONCLUSION

In this paper we have presented an image pixel based fusion technique for face recognition. After the fusion of images as weighted sum, the fused images were projected into eigenspace. Those fused eigenfaces were classified using radial basis function neural network in first case and using multilayer perceptron neural network in the second case. The efficiency of our method had been demonstrated on Object Tracking and Classification Beyond Visible spectrum (OTCBVS) benchmark database and recognition rate obtained was 96 % for RBF Neural Network and 95.07% for Multilayer Perceptron.

## ACKNOWLEDGMENT

First author is thankful to the project entitled "Development of Techniques for Human Face Based Online Authentication System Phase-I" sponsored by Department of Information Technology under the Ministry of Communications and Information

## REFERENCES

[1] P. Buddharaju, I. Pavlidis and I. Kakadiaris, "Face Recognition in the Thermal Infrared Spectrum", In Proceedings of IEEE Workshop on Computer Vision and Pattern Recognition Workshop, 2004. CVPRW '04., 27-02 June 2004.

[2] I. Pavlidis, P. Buddharaju, C. Manohar and P. Tsiamyrtzis, "Biometrics: Face Recognition in Thermal Infrared", in Biomedical Engineering Handbook, 3rd Edition, CRC Press, pp. 1-15, November, 2006.

[3] W. Zhao, R. Chellappa, A. Rosenfeld, and P. Phillips, "Face recognition: A literature survey", ACM Computer Survey, vol. 35, no. 4, pp. 399-458, December 2003.

[4] P. Jonathan Phillips, A. Martin, C. L. Wilson and M. Przybocki, "An Introduction to Evaluating Biometric System", National Institute of Standards and Technology, page 56-62, February, 2000.

[5] A. Samal and P. A. Iyenger, "Automatic recognition and analysis of human faces and facial expressions : A survey", Pattern Recognition vol. 25, No. 1, 1992.

[6] A. Gupta and S. K. Majumdar, "Machine Recognition of Human Face", http://anshulg.com/index_files/5.pdf.

[7] D. A. Socolinsky and A. Selinger, "A Comparative Analysis of Face Recognition Performance with Visible and Thermal Infrared Imagery," Proc. Int. Conf. on Pattern Recognition, Vol. 4, pp.217-222, Quebec, 2002.

[8] Z. Yin and A. A. Malcolm, "Thermal and Visual Image Processing and Fusion", http://www.simtech.a-star.edu.sg/Research/TechnicalReports.

[9] J. Heo, "Fusion of Visual and Thermal Face Recognition Techniques: A Comparative Study", The University of Tennessee, Knoxville, October 2003, http://imaging.utk.edu/publications/papers/dissertation.

[10] J. A. Taj, U. Ali, R. J. Qureshi and S. A. Khan, "Fusion of Visual and Thermal Images for Efficient Face Recognition using Gabor Filter", accepted for publication in Proc. Of The 4th ACS/IEEE International Conference on Computer Systems and Applications, March 8-11, 2006, Dubasi/Sharjah, UAE.

[11] U. Ali and J. A. Taj, "Gabor Filter based Efficient Thermal and Visual Face Recognition Fusion Architecture" accepted for publication in the fourth International Conference on Active Media Technology, June 7-9 2006 Brisbane, Australia.

[12] David Hughes, "Sinking in a Sea of Pixels- The Case for Pixel Fusion", 2006 Silicon Graphics, Inc., http://www.sgi.com/pdfs.

[13] M. Turk and A. Pentland, "Eigenfaces for recognition", Journal of Cognitive Neuro-science, March 1991. Vol 3, No-1, pp-71-86.

[14] L. Sirovich and M. Kirby, "A low-dimensional procedure for the characterization of human faces," J. Opt. Soc. Amer. A 4(3), pp. 519-524, 1987.

[15] R. Gottumukkal and V. K. Asari, "An improved face recognition technique based on modular PCA approach", Pattern Recognition Letters 25:429-436, 2004.

[16] Er. M. Joo, S. Wu, J. Lu and H. L. Toh, "Face Recognition With Radial Basis Function (RBF) Neural Networks", In Proceedings of IEEE Transactions of Neural Networks, Vol. 13, No. 3, May 2002.

[17] J. Haddadnia, K. Faez and M. Ahmadi, "An Efficient Human Face Recognition System using Pseudo Zernike Moment invariant and radial Basis Function Neural Network", In International Journal of Pattern Recognition and Artificial Intelligence, Vol. 17, No. 1, (2003).

[18] J. Moody and C. J. Darken, "Fast learning in network of locally-tuned processing units," Neural Comput., vol. 1, pp. 281–294, 1989.

[19] S. Lee and R. M. Kil, "A Gaussian potential function network with hierarchically self-organizing learning," Neural Networks, vol. 4, pp. 207–224, 1991.

[20] W. Pedrycz, "Conditional fuzzy clustering in the design of radial basis function neural networks," IEEE Trans. Neural Networks, vol. 9, pp. 601–612, July 1998.

## REFERENCES

[1] P. Buddharaju, I. Pavlidis and I. Kakadiaris, "Face Recognition in the Thermal Infrared Spectrum", In Proceedings of IEEE Workshop on Computer Vision and Pattern Recognition Workshop, 2004. CVPRW '04., 27-02 June 2004.

[2] I. Pavlidis, P. Buddharaju, C. Manohar and P. Tsiamyrtzis, "Biometrics: Face Recognition in Thermal Infrared", in Biomedical Engineering Handbook, 3rd Edition, CRC Press, pp. 1-15, November, 2006.

[3] W. Zhao, R. Chellappa, A. Rosenfeld, and P. Phillips, "Face recognition: A literature survey", ACM Computer Survey, vol. 35, no. 4, pp. 399-458, December 2003.

[4] P. Jonathan Phillips, A. Martin, C. L. Wilson and M. Przybocki, "An Introduction to Evaluating Biometric System", National Institute of Standards and Technology, page 56-62, February, 2000.

[5] A. Samal and P. A. Iyenger, "Automatic recognition and analysis of human faces and facial expressions : A survey", Pattern Recognition vol. 25, No. 1, 1992.

[6] A. Gupta and S. K. Majumdar, "Machine Recognition of Human Face", http://anshulg.com/index_files/5.pdf.

[7] D. A. Socolinsky and A. Selinger, "A Comparative Analysis of Face Recognition Performance with Visible and Thermal Infrared Imagery," Proc. Int. Conf. on Pattern Recognition, Vol. 4, pp.217-222, Quebec, 2002.

[8] Z. Yin and A. A. Malcolm, "Thermal and Visual Image Processing and Fusion", http://www.simtech.a-star.edu.sg/Research/TechnicalReports.

[9] J. Heo, "Fusion of Visual and Thermal Face Recognition Techniques: A Comparative Study", The University of Tennessee, Knoxville, October 2003, http://imaging.utk.edu/publications/papers/dissertation.

[10] J. A. Taj, U. Ali, R. J. Qureshi and S. A. Khan, "Fusion of Visual and Thermal Images for Efficient Face Recognition using Gabor Filter", accepted for publication in Proc. Of The 4th ACS/IEEE International Conference on Computer Systems and Applications, March 8-11, 2006, Dubasi/Sharjah, UAE.

[11] U. Ali and J. A. Taj, "Gabor Filter based Efficient Thermal and Visual Face Recognition Fusion Architecture" accepted for publication in the fourth International Conference on Active Media Technology, June 7-9 2006 Brisbane, Australia.

[12] David Hughes, "Sinking in a Sea of Pixels- The Case for Pixel Fusion", 2006 Silicon Graphics, Inc., http://www.sgi.com/pdfs.

[13] M. Turk and A. Pentland, "Eigenfaces for recognition", Journal of Cognitive Neuro-science, March 1991. Vol 3, No-1, pp-71-86.

[14] L. Sirovich and M. Kirby, "A low-dimensional procedure for the characterization of human faces," J. Opt. Soc. Amer. A 4(3), pp. 519-524, 1987.

[15] R. Gottumukkal and V. K. Asari, "An improved face recognition technique based on modular PCA approach", Pattern Recognition Letters 25:429-436, 2004.

[16] Er. M. Joo, S. Wu, J. Lu and H. L. Toh, "Face Recognition With Radial Basis Function (RBF) Neural Networks", In Proceedings of IEEE Transactions of Neural Networks, Vol. 13, No. 3, May 2002.

[17] J. Haddadnia, K. Faez and M. Ahmadi, "An Efficient Human Face Recognition System using Pseudo Zernike Moment invariant and radial Basis Function Neural Network", In International Journal of Pattern Recognition and Artificial Intelligence, Vol. 17, No. 1, (2003).

[18] J. Moody and C. J. Darken, "Fast learning in network of locally-tuned processing units," Neural Comput., vol. 1, pp. 281–294, 1989.

[19] S. Lee and R. M. Kil, "A Gaussian potential function network with hierarchically self-organizing learning," Neural Networks, vol. 4, pp. 207–224, 1991.

[20] W. Pedrycz, "Conditional fuzzy clustering in the design of radial basis function neural networks," IEEE Trans. Neural Networks, vol. 9, pp. 601–612, July 1998.